\documentclass{article}

\usepackage{ijcai16}

\usepackage{times}
\usepackage{epsfig}
\usepackage{graphicx}
\usepackage{amsmath}
\usepackage{amssymb}
\usepackage{mathsymb}
\usepackage{textcomp}
\usepackage[linesnumbered,algoruled,boxed,lined]{algorithm2e}
\usepackage{verbatim}
\usepackage{subfigure}
\usepackage{url}
\usepackage{graphicx}
\usepackage{multirow}
\usepackage[super]{nth}
\usepackage{balance}
\usepackage{colortbl}
\graphicspath{{./}{./Fig/}{./Figs/}{./Fig/eps/}{./Fig/pdf/}}

\usepackage{color}
\usepackage{xcolor}
\usepackage{algorithm2e}

\let\savedalgorithm\algorithm
\let\savedendalgorithm\endalgorithm

\usepackage{algorithm}

\newcommand{\eg}{\emph{e.g.,\ }}
\newcommand{\ie}{\emph{i.e.,\ }}

\title{Crossing Generative Adversarial Networks for Cross-View Person Re-identification}
\author{Chengyuan Zhang$^{\natural}$, Lin Wu$^{\ddag}$, Yang Wang$^{\dag}$\\
 $^{\natural}$School of Information Science and Engineering, Central South University, Changsha 410083, China\\
 $^{\ddag}$ISSR, ITEE, The University of Queensland, Brisbane, QLD, 4072, Australia\\
 $^{\dag}$The University of New South Wales, Kensington, Sydney, Australia\\
 Correspondence to lin.wu@uq.edu.au
}

\begin{document}

\maketitle

\begin{abstract}

Person re-identification (\textit{re-id}) refers to matching pedestrians across disjoint yet non-overlapping camera views. The most effective way to match these pedestrians undertaking significant visual variations is to seek reliably invariant features that can describe the person of interest faithfully. Most of existing methods are presented in a supervised manner to produce discriminative features by relying on labeled paired images in correspondence. However, annotating pair-wise images is prohibitively expensive in labors, and thus not practical in large-scale networked cameras. Moreover, seeking comparable representations across camera views demands a flexible model to address the complex distributions of images. In this work, we study the co-occurrence statistic patterns between pairs of images, and propose to crossing Generative Adversarial Network (Cross-GAN) for learning a joint distribution for cross-image representations in a unsupervised manner. Given a pair of person images, the proposed model consists of the variational auto-encoder to encode the pair into respective latent variables, a proposed cross-view alignment to reduce the view disparity, and an adversarial layer to seek the joint distribution of latent representations. The learned latent representations are well-aligned to reflect the co-occurrence patterns of paired images. We empirically evaluate the proposed model against challenging datasets, and our results show the importance of joint invariant features in improving matching rates of person re-id with comparison to semi/unsupervised state-of-the-arts.

\end{abstract}

\section{Introduction}\label{sec:intro}

Nowadays person re-identification (re-id) is emerging as a key problem in intelligent surveillance system, which deals with maintaining identities of individuals at physically different locations through non-overlapping camera views. Cross-view person re-id enables automated discovery and analysis of person specific long-term structural activities over wide areas, and is fundamental to many surveillance applications such as multi-camera people tracking and forensic search.

More recently, deep learning methods gradually gain the popularity in person re-id, which are developed to incorporate two aspects of feature extraction and metric learning into an integrated framework \cite{FPNN,JointRe-id,SI-CI,DomainDropout,DeepRanking,PersonNet,DeepReID,GatedCNN}. The basic idea is to feed-forward a pair of input images into two CNNs with shared weights to extract features, and a subsequent metric learning part compares the features to measure the similarity. This process is carried out essentially by a classification on \textit{cross-image representation} whereby images are coupled to extract their features, after which a parameterized classifier based on some distance measure (\eg Euclidean distance) performs an ordinary binary classification task to predict whether the two pedestrian images are from the same person. The cross-image representation is effective in capturing the relationship across pairs of images, and several approaches have been suggested to address horizontal displacement by local patch matching. For instance, the FPNN \cite{FPNN} algorithm introduced a patch matching layer for the CNN part at early layers. An improved deep learning architecture is proposed in \cite{JointRe-id} with cross-input neighborhood differences and patch summary features.These two methods are both dedicated to improve the CNN architecture with a purpose to evaluate the pair similarity early in the CNN stage, so that it could make use of spatial correspondence of feature maps. Adding on, in \cite{GatedCNN}, a matching gate is embedded into CNN to extract more locally similar patterns in horizontal correspondence across viewpoints. As for the metric learning part, with the aim to reduce the distance of matched images while enlarging the distance of mismatched images, common choices are pairwise and/or triplet comparison constraints. For example, \cite{FPNN,JointRe-id,PersonNet} use the logistic loss to directly form a binary classification problem of whether the input image pair belongs to the same identity. In some other works, \cite{GatedCNN} adopts the contrastive loss based on pairwise comparison. \cite{DeepRanking} uses Euclidean distance and triplet loss while \cite{SI-CI} optimizes the combination loss function based on pairwise and triplet constraints.

\begin{figure}[t]
\centering
\includegraphics[height=3cm]{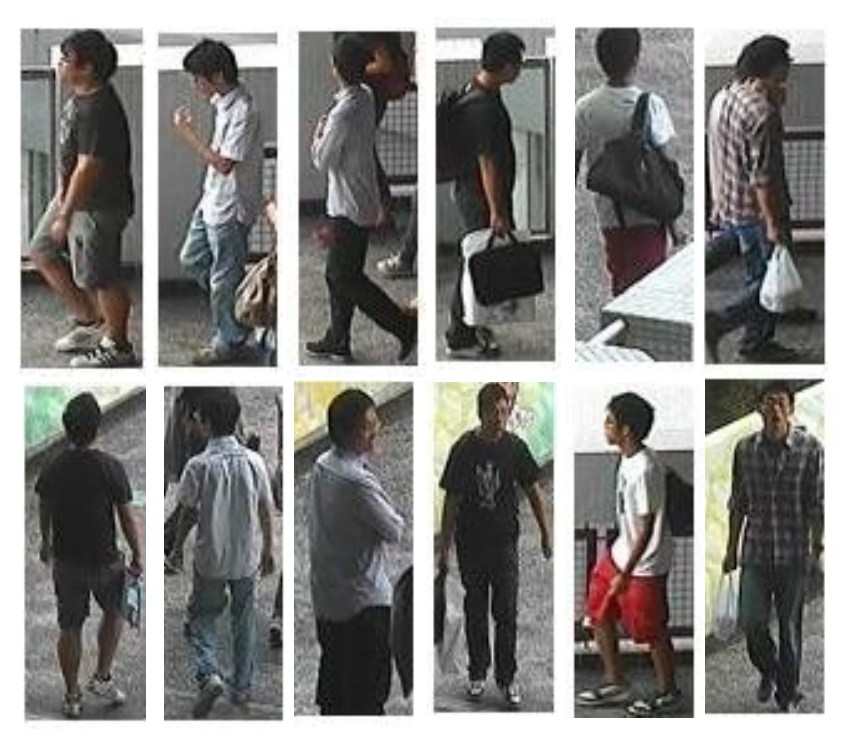}
\includegraphics[height=3cm]{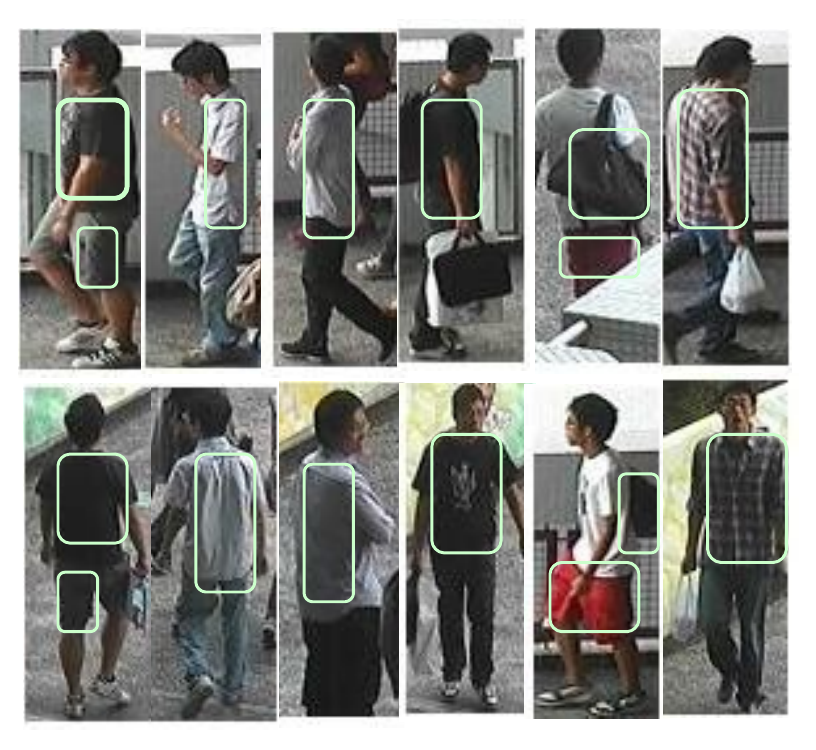}
\caption{Left: Pedestrian images selected from CUHK03 dataset. Each column indicates images in pairs regarding the same person observed by disjoint camera views. Right: Illustration of co-occurrence regions in positive image pairs.}
\label{fig:example}
\end{figure}

However, these deep learning methods are inherently limited due to two presumable assumptions: the availability of large numbered labeled samples across views and the two fixed camera views are supposed to exhibit a unimodal inter-camera transform. In practice, building a training dataset with tuples of labeled corresponding images is impossible for every pair of camera views in the context of a large camera network in video surveillance. Thus, this correspondence dependency greatly limits the applicability of the existing approaches with training samples in correspondence. Secondly, the practical configurations (which are the combinations of view points, poses, lightings, and photometric settings) of pedestrian images are \textit{multi-modal} and view-specific \cite{LocallyAlligned} even if they are observed under the same camera. Therefore, the complex yet multi-modal inter-camera variations cannot be well learned with a generic metric  which is incapable of handling multiple types of transforms across views. Last but not the least, existing deep learning methodologies directly compute the difference between intermediate CNN features and propagate only distance/similarity value to a ultimate scalar. This would lose important information since they did not consider feature alignment in cross-view.

\subsection{Our Approach and Contributions}
\begin{figure}[t]
\centering
\includegraphics[height=3cm]{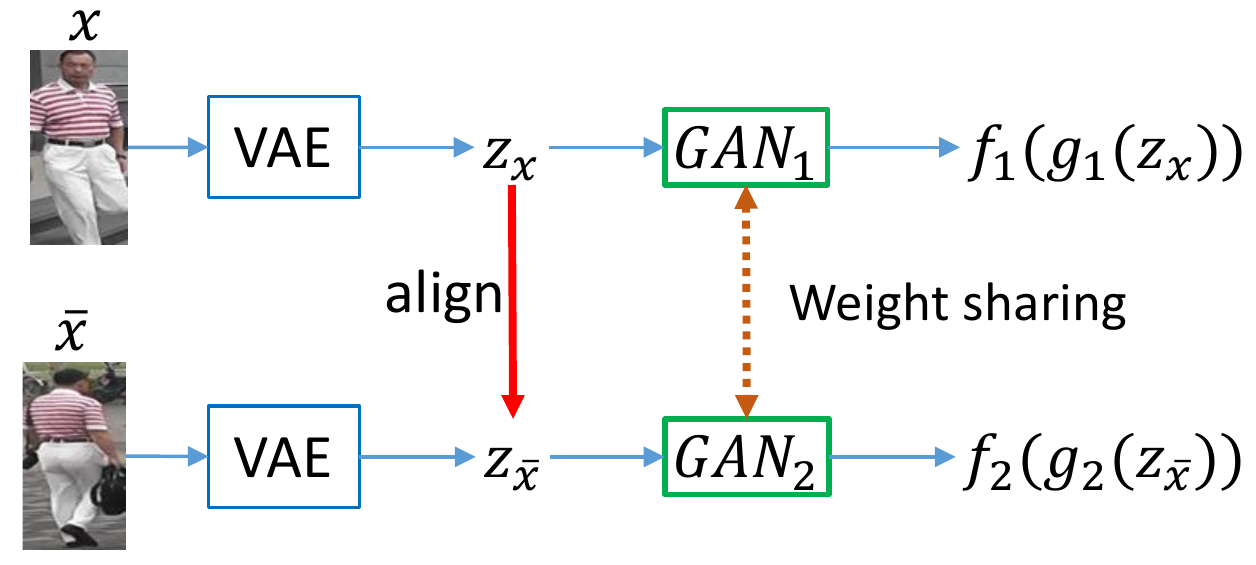}
\caption{The schematic overview of the proposed crossing GAN for person re-id.}
\label{fig:simple_VAE_GAN}
\end{figure}

To overcome these limitations, we propose the crossing net based on a couple of generative adversarial networks (GANs) \cite{GAN} to seek effective cross-view representations for person re-id. To combat the first issue of relying on supervision, as shown in Fig.\ref{fig:example}, we observe some patterns that appear commonly across image pairs are distinct to discriminate positive pairs from negatives. Thus, these co-occurrence patterns should be mined out automatically to facilitate the task of re-id. Specifically, as shown in Fig.\ref{fig:simple_VAE_GAN}, the proposed network starts from a tuple of variational auto-encoder (VAE) \cite{AEV}, each for one image from a camera view, to encode the input images into their respective latent variables without any region-level annotations on person images. The technique of VAE has been established a viable solution for image distribution learning tasks while in this paper, we employ VAE to statistically generate latent variables for paired images without correspondence labeling. We remark that we don't use the Siamese Convolutional Neural Networks (CNNs) \cite{GatedCNN} to encode the input pair because CNNs are composed of fixed receptive fields which may not flexible to capture the varied local patterns. Also, the Siamese architecture enforces the weight sharing across CNN layers which are not suited for multi-modal view-specific variations.

To address the view disparity, we propose a cross-view alignment which is bridged over VAE outputs to allow the comparable matching. This alignment operation is to derive a shared latent space by modeling the statistical relationships between generative variables, and we empirically demonstrate this explicit alignment is crucial for cross-view representation learning (see Section \ref{ssec:ablation}). Then, the crossing net is coupled with adversarial networks to produce joint view-invariant distribution which gives a probability function to each joint occurrence of cross-view person images.

The major contributions of this paper can be summarized as follows:
\begin{itemize}
\item We extend the GAN to a dual setting, namely Cross-GAN, which is augmented with VAE to learn jointly invariant features for the task of person re-id in a unsupervised manner.
\item The proposed Cross-GAN consists of a VAE layer to effectively encode image distributions w.r.t each camera view, a view-alignment layer to discover a shared latent space between cross-view images, and an adversarial network to produce the joint distribution of images.
\item Extensive experiments are conducted to demonstrate our method outperforms semi/unsupervised state-of-the-art yet very comparable to supervised methods.
\end{itemize}

\section{Related Work}\label{sec:related}

\subsection{Person Re-identification}

The task of person re-identification can be accomplished by two categories of methods: (i) learning distance or similarity measures to predict if two images describe the same person \cite{LADF,Xiong2014Person,LocallyAlligned,Zheng2011Person,NullSpace-Reid,YangMM13,YangMM14,YangCIKM13,SimilaritySpatial,YangNN18,YangTNNLS18,LocalMetric,E-Metric}, and (ii) designing distinctive signature to represent a person under different cameras, which typically performs classification on cross-image representation \cite{FPNN,JointRe-id,GatedCNN,SI-CI,Deep-Embed,SimilaritySpatial}. For the first category of methodologies, they usually use many kinds of hand-crafted features including local binary patterns \cite{Xiong2014Person,KISSME,YangMM15,YangCVIU17,YangINS13,YangNeurocomputing13,Yangpakdd14,YangKAIS16}, color histogram \cite{KISSME,YangIVC17,LDAFisherVector,YangTcyb17}, local maximal occurrence (LOMO) \cite{LOMOMetric,PSD-metric}, and focus on learning an effective distance/similarity metric to compare the features. For the second category, deep convolutional neural networks are very effective in localizing/extracting relevant features to form discriminative representations against view variations. However, all these re-id models are in a supervised manner and rely on substantial labeled training data, which are typically required to be in pair-wise for each pair of camera views. Their performance depends highly on the quantity and quality of labeled training data, which also limits their application to large-scale networked cameras. In contrast, our method is based on unsupervised generative modeling which does not require any labeled data, and thus is free from prohibitively high cost of manual labeling and the risk of incorrect labeling.

A body of unsupervised methods have been developed to address person re-id without dependency on labeling \cite{LOMOMetric,eSDC,CAMEL,Farenzena2010Person,YangSIGIR15,YangIJCAI162,YangTNNLS172,OL-MANS,UMDL,LinPR172,YangTIP17,One-shot-RE-ID,YangTIP15}. These models differ from ours in two aspects. On the one hand, these models do not explicitly model the view-specific information, i.e., they treat feature transformation/optimization in every distinct camera view in the same manner. In contrast, our models is propertied to employ VAE to generate view-specific latent variables, and then aim to find a shared subspace through a view-alignment layer. Thus, view-specific interference can be alleviated and common patterns can be attained in the representation learning. On the other hand, our method is the first attempt to introduce the adversarial learning into cross-view representation learning which can automatically discover co-occurrence patterns across images. While co-occurrence based statistics has been studied in some work \cite{Co-occurrence,Co-occurrence-object,Co-occurrence-graph,LOMOMetric}, our approach diverts from the literature by aiming to jointly optimized invariant feature distributions for cross-image representations.

\subsection{Deep Generative Models}

In recent years, generative models have received an increasing amount of attention. Several approaches including variational auto-encoders (VAE) \cite{AEV,Stochastic-BPP-ICML2014}, generative adversarial networks (GAN) \cite{GAN}, and attention models \cite{DRAW} have shown that learned deep networks are capable of generating new data points after the completion of training to learn an image distribution from unlabeled samples. Typically, determining the underlying data distribution of unlabeled images can be highly challenging and inference on such distributions is highly computationally expensive and or intractable except in the simplest of cases. VAE and GAN are the most prominent ones which provide efficient approximations, making it possible to learn tractable generative models of unlabeled images.

Our proposed network is inspired by the coupled generative adversarial networks \cite{CoGAN}, which learn a joint distribution of images without any tuple of corresponding images. It is demonstrated to be applied into domain adaptation and image transformation. Whilst our method has the sharing of coupled GANs in terms of enforcing weight sharing across the streamed GANs, our model is different from \cite{CoGAN} on two facets. First, the model of \cite{CoGAN} is originated from the same source of random vector as the uniform distribution for the generator of GANs whereas our method uses two respective VAE to generate the random vectors for two GANs. Second, our model has a cross-view alignment layer to seek a shared latent space for two distributions which is not provided in \cite{CoGAN}.

\section{Preliminaries}\label{sec:preliminary}
Let $\bx$ and $\bar{\bx}$ represent a pair of observations (\eg two images of pedestrians). We aim to learn a set of latent random variables $\bz$ and $\bar{\bz}$ ($\bz$ and $\bar{\bz}$ are linked by an alignment mapping), designed to capture the variations in the observed inputs while maintaining co-occurrence therein. To this end, we wish to estimate a prior $p(\bx)$ ($p(\bar{\bx})$) by modeling the generation process of $\bx$ ($\bar{\bx}$) by sampling some $\bz$ ($\bar{\bz}$) from an arbitrary distribution $p(\bz)$ ($p(\bar{\bz})$) as $p(\bx)=\int_{\bz} p(\bx|\bz) p(\bz) d\bz$ ($p(\bar{\bx})=\int_{\bar{\bz}} p(\bar{\bx}|\bar{\bz}) p(\bar{\bz}) d\bar{\bz}$). Fitting $p(\bx)$ ($p(\bar{\bx})$) directly is intractable which involves expensive inference. We therefore approximate $p(\bx)$ and $p(\bar{\bx})$ using VAE on each, respectively, because VAE offers a combination of highly flexible non-linear mapping between the latent states and the observed output and effective approximate inference. To further induce joint invariant distribution between $\bz$ and $\bar{\bz}$, two respective VAEs are connected with two GANs through which the shared latent representations to images in individual can be attained by an adversary acting on pairs of $(\bx,\bar{\bx})$ data points and their latent codes $(\bz,\bar{\bz})$.
In the remainder of this section, we provide brief introduction of VAE and GAN which we use to model the prior of pedestrian images and joint invariant distributions.

\subsection{Variational Autoencoder (VAE)}

A VAE comprises an encoder which estimates the posterior of latent variable and a decoder generates sample from latent variable as follows,
\begin{equation}
\bz \sim \mbox{encoder}(\bx)=q(\bz|\bx), \hat{\bx} \sim \mbox{decoder}(\bz_{\bx})=p(\bx|\bz).
\end{equation}
The VAE regularizes the encoder by imposing a prior over the latent distribution on $p(\bz)$ while at the same time reconstructing $\hat{\bx}$ to be as close as possible to the original $\bx$. Typically, $q(\bz|\bx)$ is taken to be a Gaussian prior, \ie $\bz\sim \mathcal{N}(0,1)$, which can be incorporated into a loss in the form of Kullback-Leibler divergence $D_{KL}$ between the encoded distribution $q(\bz|\bx)$ and the prior $p(\bz)$. Thus, the VAE loss takes the form of the sum of the reconstruction error and latent prior:
\begin{equation}\label{eq:vae_loss}
\mathcal{L}_{vae}=D_{KL} \left(q(\bz|\bx) || p(\bz) \right) - \mathbb{E}_{q(\bz|\bx)}[\log p(\bx | \bz)].
\end{equation}
We use the VAE to be an effective modelling paradigm to recover the complex multi-modal distributions of images over the data space.  A VAE introduces a set of latent random variables $\bz$, designed to capture the variations in the observed variable $\bx$.

\subsection{Generative Adversarial Networks (GAN)}

A GAN consists of a generator and a discriminator. The objective of the generator is to synthesize images resembling real images, while the objective of the discriminator is to distinguish real images from synthesized ones. Let $\bx$ be a natural image drawn from distribution $p_X$, and $\bz$ be a random vector in $\mathbb{R}^d$. Let $g$ and $f$ be the generative and discriminative models, respectively. The generator synthesizes samples by mapping a random sample $\bz$, from an arbitrary distribution, to a sample as output image $g(\bz)$, that has the same vector support as $\bx$. Denote the distribution of $g(\bz)$ as $p_G$. The discriminator tries to distinguish between real data sample $\bx$, and synthesized sample $g(\bz)$ by estimating the probability that an input image is drawn from $p_X$. The loss function for the GAN can be formulated as a binary entropy loss as follows:
\begin{equation}\label{eq:GAN}
\mathcal{L}_{gan} (f,g)= \log f(\bx)+ \log (1-f(g(\bz))).
\end{equation}
Training on Eq.\eqref{eq:GAN} alternatives between minimizing $\mathcal{L}_{gan}$ w.r.t. parameters of the generator while maximizing $\mathcal{L}_{gan}$ w.r.t. parameters of the discriminator. The generator tries to minimize the loss to generate more realistic samples to fool the discriminator while the discriminator tries to maximize the loss.

In practice, Eq.\eqref{eq:GAN} is solved by alternating the following gradient update steps:
\begin{itemize}
\item $\boldsymbol \theta_f^{t+1}= \boldsymbol \theta_f^t - \lambda^t \nabla_{\boldsymbol \theta_f} \mathcal{L}_{gan}(f^t, g^t)$,
\item $\boldsymbol \theta_g^{t+1}= \boldsymbol \theta_g^t - \lambda^t \nabla_{\boldsymbol \theta_g} \mathcal{L}_{gan}(f^{t+1}, g^t)$.
\end{itemize}
where $\boldsymbol \theta_f$ and $\boldsymbol \theta_g$ are parameters of $f$ and $g$, $\lambda$ is the learning rate, and $t$ is the iteration number. The GAN does not explicitly model reconstruction loss of the generator; instead, network parameters are updated by back-propagating gradients only from the discriminator. This strategy can effectively avoid pixel-wise loss functions that tend to produce overly smoothed results and enables realistic modeling of noise as present in the training set. Thus, GAN can be used to synthesize images, i.e., the distribution $p_G$ converges to $p_X$, given enough capacity $f$ and $g$ and sufficient training iterations \cite{GAN}.

\section{The Method}\label{sec:CrossGAN}

\begin{figure*}[!ht]
\centering
\includegraphics[height=5.5cm]{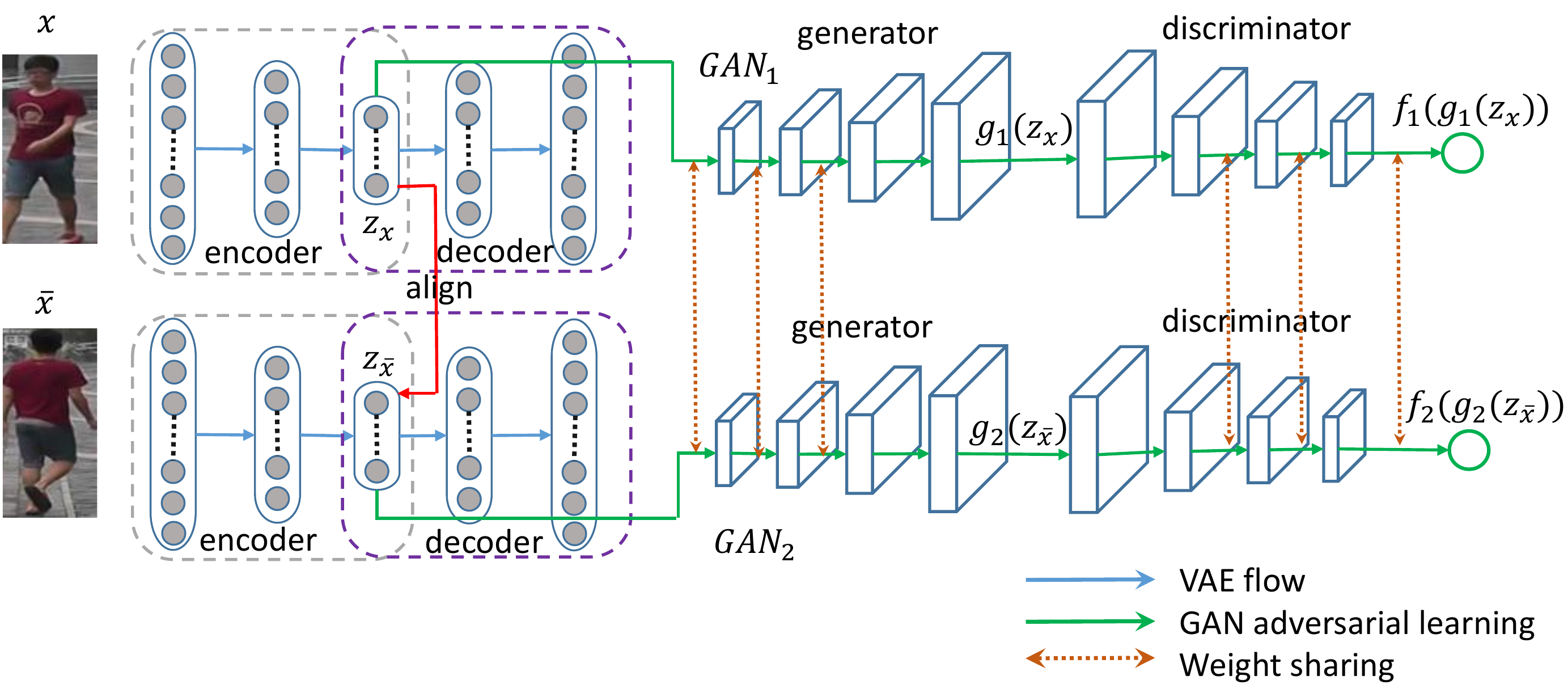}
\caption{Architecture overview. Best view in color.}
\label{fig:system}
\end{figure*}

\subsection{System Overview: Crossing GANs}

The complete network is then trained end-to-end for learning a joint invariant distribution of images across camera views. Fig.\ref{fig:system} illustrates the overview of our architecture. It consists of a pair of (VAE, GAN)s, that is, $(VAE_1, GAN_1)$ and $(VAE_2, GAN_2)$; each is responsible for synthesizing one image in one camera view. In Fig.\ref{fig:system}, the blue and green routes represent the forward paths of the VAE and GAN for images $\bx$ and $\bar{\bx}$, respectively. The blue route, \ie the VAE flow, is the use of expressive latent variables to model the variability observed in the data. It essentially captures the statistics of each individual image. The auto-encoding procedure is explained in Section \ref{ssec:encoding}. The red route denotes the cross-view alignment that links the latent variables ($\bz_{\bx},\bz_{\bar{\bx}}$) to ensure the shared latent representations. The details of alignment is given in Section \ref{ssec:alignment}. The green routes represent the adversarial learning which works to optimize optimal latent features corresponding to the joint invariance across paired images. During training, the two GANs are enforced to share a subset of parameters (the brown routes), which results in synthesized pairs of corresponding images without correspondence supervision. The details are described in Section \ref{ssec:adversarial}.

\subsection{Auto-encoding}\label{ssec:encoding}

Given a pair of data points ($\bx^{(i)}, \bar{\bx}^{(i)}$) from a dataset $\bX=\{\bx^{(i)}, \bar{\bx}^{(i)}\}_{i=1}^M$ containing $N=2M$ samples in $M$ pairs.
The auto-encoding algorithm uses unobserved random variable $\bz^{(i)}$, to generate a data point $\bx^{(i)}$. As the generating process can be repeated on either $\bx^{(i)}$ or $\bar{\bx}^{(i)}$, in the following, we describe $\bx^{(i)}$ as illustration. The process is composed of two phases: (1) a value $\bz^{(i)}$ is generated from some prior distribution $p(\bz^{(i)})$; (2) a value $\bx^{(i)}$ is generated from some conditional distribution $p(\bx^{(i)} | \bz^{(i)})$. From a coding theory perspective, the unobserved variable $\bz^{(i)}$ have an interpretation as a latent representation or \textit{code}. Following VAE \cite{AEV} which introduces a recognition model $q(\bz^{(i)}|\bx^{(i)})$: an approximation to the intractable true posterior $p(\bz^{(i)}|\bx^{(i)})$, we will therefore refer to $q(\bz^{(i)}|\bx^{(i)})$ as a probabilistic \textit{encoder}, since given a data point $\bx^{(i)}$ its produces a distribution (\eg a Gaussian) over the possible values of the code $\bz^{(i)}$ from which the data point $\bx^{(i)}$ could be generated. In a similar vein, we refer to $p(\bx^{(i)}|\bz^{(i)})$ as a probabilistic \textit{decoder}, since given a code $\bz^{(i)}$ it produces a distribution over the possible corresponding values of $\bx^{(i)}$.

In this work, neural networks are used as probabilistic encoders and decoders, namely multi-layered perceptions (MLPs). Let the prior over the latent variables be the centered isotropic multivariate Gaussian $p(\bz)=\mathcal{N}(\bz; \boldsymbol 0, \boldsymbol I)$ whose distribution parameters are computed from $\bz$ with a MLP. We assume the true posterior $p(\bz|\bx)$ takes on an approximate Gaussian form with an approximately diagonal covariance. In this case, we can let the variational approximate posterior be a multivariate Gaussian with a diagonal covariance structure:
\begin{equation}
\log q(\bz^{(i)}|\bx^{(i)})=\log \mathcal{N}(\bz; \boldsymbol\mu^{(i)}, \boldsymbol\sigma^{2(i)} \boldsymbol I)
\end{equation}
where the mean and standard of the approximate posterior, $\boldsymbol\mu^{(i)}, \boldsymbol\sigma^{(i)}$ are outputs of the encoding MLP. \ie nonlinear functions of data point $\bx^{(i)}$ and the variational parameters.

Specifically, we sample from the posterior $\bz^{(i)}\sim q(\bz|\bx^{(i)})$ using $\bz^{(i)}=\boldsymbol\mu^{(i)} + \boldsymbol\sigma^{(i)} \odot \boldsymbol\epsilon$ where $\boldsymbol\epsilon \sim \mathcal{N}(\boldsymbol 0, \boldsymbol I)$. With $\odot$ we signify an element-wise product. In this model, both $p(\bz)$ and $q(\bz|\bx)$ are Gaussian. The resulting estimator loss for data point $\bx^{(i)}$ is:
\begin{equation}\label{eq:estimator}\small
\begin{split}
&\tilde{L}_{vae}(\bx^{(i)})= D_{KL}(q(\bz|\bx^{(i)}|| p(\bz))) - \mathbb{E}_{q(\bz|\bx^{(i)})} [\log p(\bx^{(i)}|\bz)]\\
&=D_{KL}(q(\bz|\bx^{(i)}|| p(\bz))) -\log p(\bx^{(i)}|\bz^{(i)})\\
&\simeq \frac{1}{2}\sum_{j=1}^J \left(1+\log((\sigma_j^{(i)})^2)-(\mu_j^{(i)})^2- (\sigma_j^{(i)})^2\right) - \log p(\bx^{(i)}|\bz^{(i)})\\
& \bz^{(i)}=\boldsymbol\mu^{(i)} + \boldsymbol\sigma^{(i)} \odot \boldsymbol\epsilon, \boldsymbol\epsilon \sim \mathcal{N}(\boldsymbol 0, \boldsymbol I)
\end{split}
\end{equation}
where the KL-divergence $D_{KL}(q(\bz|\bx^{(i)}|| p(\bz)))$ can be integrated analytically, such that only the expected reconstruction error $\mathbb{E}_{q(\bz|\bx^{(i)})} [\log p(\bx^{(i)}|\bz)]$ requires estimation by sampling. Given multiple data points from a dataset $\bX$ with $M$ pairs of data points, we can construct an estimator loss of as follows:
\begin{equation}
\tilde{L}_{vae}(\bX)=\frac{1}{M}\sum_{i=1}^M \left(\tilde{L}_{vae}(\bx^{(i)}) + \tilde{L}_{vae}(\bar{\bx}^{(i)}) \right).
\end{equation}

\subsection{Learning Cross-View Alignment on Latent Codes}\label{ssec:alignment}

In this section, we introduce cross-view alignment over latent representations provided by VAE, which is capable of modeling complex multi-modal distributions over data space.
Note that for notation convenience, we use $\bz_{\bx}$ and $\bz_{\bar{\bx}}$ to distinguish the latent representation for $\bx$ and $\bar{\bx}$.

\begin{equation}\label{eq:align}
\mathcal{L}_{align}=\max(|| \bz_{\bx}-\mbox{ Align}(\bz_{\bar{\bx}}) ||^2, \tau),
\end{equation}
where we model $\mbox{Align}(\cdot)$ as a single fully connected neuron with a $tanh$ activation function. The threshold $\tau$ is $\tau=1$. In essence, $\mbox{Align}(\cdot)$ is implicitly learning a mapping across two normal distributions ($\bz_{\bx},\bz_{\bar{\bx}}$). The parameters of the mapping $\boldsymbol\theta_{Align}$ are optimized through back-propagation. Since both the VAE and the GAN are able to learn low-dimensional representations (in our case, both $\bz_{\bx},\bz_{\bar{\bx}}$ are set to be 100 dimensions.), we are able to fit the cross-view alignment with moderate pairs.

The strategy of alignment is designed to align the transformation across cameras by revealing underlying invariant properties among different views. As a result, unsupervised matching pedestrian images can be statistically inferred through aligned latent representations. This is motivated by the observation that some regions are distributed similarly in images across views and robustly maintain their appearance in the presence of large cross-view variations.

\subsection{Adversarial Learning}\label{ssec:adversarial}

\subsubsection{Generator}
Let $g_1$ and $g_2$ be the generators of $GAN_1$ and $GAN_2$, which map corresponding inputs $ \bz_{\bx}$ and $\bz_{\bar{\bx}}$ to images that have the same support as $\bx$ and $\bar{\bx}$, respectively. Both $g_1$ and $g_2$ are realized as convolutions \cite{ConvGAN}:
\begin{equation}
\begin{split}
& g_1(\bz_{\bx})=g_1^{(m)} (g_1^{(m-1)} (\ldots g_1^{(2)} (g_1^{(1)}(\bz_{\bx}) ))),\\
& g_2(\bz_{\bar{\bx}})=g_2^{(m)} (g_2^{(m-1)} (\ldots g_2^{(2)} (g_2^{(1)}(\bz_{\bar{\bx}}) )));
\end{split}
\end{equation}
where $g_1^{(i)}$ and $g_2^{(i)}$ are the $i$-th layer of $g_1$ and $g_2$ and $m$ is the number of layers in generators. Through layers of convolution operations, the generator gradually decode information from more abstract concept to more material details. The first layer decode high-level semantics while the last layer decode low-level details. Note this information flow is opposite to that in a standard deep neural network \cite{AlexNet} where the first layers extract low-level features while the last layers extract high-level features. Based on the observation that a pair of person images from two camera views share the same high-level concept (i.e., they belong to the same identity but with different visual appearance), we enforce the first layers of $g_1$ and $g_2$ to have identical structures and share the weights, which means $\boldsymbol \theta_{g_1^{(i)}}=\boldsymbol \theta_{g_2^{(i)}}$, for $i=1,2,\ldots,k$ where $k$ is the number of shared layers, and $\theta_{g_1^{(i)}}$ and $\theta_{g_2^{(i)}}$ are the parameters of $g_1^{(i)}$ and $g_2^{(i)}$, respectively. This constraint can force the high-level semantics to be decoded in the same way in $g_1$ and $g_2$, which can also be propagated into the VAE to update the parameters simultaneously. Thus, the generator can gradually decode the information from more abstract concepts to more finer details, and the view-alignment is embedded to ensure the common finer regions can be preserved with high correlations.

\subsubsection{Discriminator}

Let $f_1$ and $f_2$ be the discriminators of $GAN_1$ and $GAN_2$ given by
\begin{equation}
\begin{split}
& f_1(\bx)=f_1^{(n)} (f_1^{(n-1)} (\ldots f_1^{(2)} (f_1^{(1)}(\bx) ))),\\
& f_2(\bar{\bx})=f_2^{(n)} (f_2^{(n-1)} (\ldots f_2^{(2)} (f_2^{(1)}(\bar{\bx}) )));
\end{split}
\end{equation}
where $f_1^{(i)}$ and $f_2^{(i)}$ are the $i$-th layer of $f_1$ and $f_2$, and $n$ is the number of layers. Note that $GAN_1$ and $GAN_2$ have the identical network structure.  The discriminator maps an input image to a probability score, estimating the likelihood that the input is drawn from a true data distribution. The first layers of the discriminator extract low-level features while the last layers of layers extract high-level features. Considering that input image pair are realizations of the same person in two camera views, we force $f_1$ and $f_2$ to have the same last layers, which is achieved by sharing the weights of the last layers via $\boldsymbol \theta_{f_1^{(n-i)}}=\boldsymbol \theta_{f_2^{(n-i)}}$, for $i=0,1,\ldots,l-1$ where $l$ is the number of weight-sharing layers in the discriminator, and $\boldsymbol \theta_{f_1^{(i)}}$ and $\boldsymbol \theta_{f_2^{(i)}}$ are the network parameters of $f_1^{(i)}$ and $f_2^{(i)}$, respectively. The weight-sharing constraints herein helps reduce the number of trainable parameters of the network, and also effective in deriving view-invariant features in joint distribution across $\bx$ and $\bar{\bx}$.

Therefore, we cast the problem of learning jointly invariant feature distribution as a constrained objective function with the training loss given by
\begin{equation}\label{eq:cross-loss}
\begin{split}
\mathcal{L}_{gan}(f_1,f_2,g_1,g_2)=\log f_1(\bx)+ \log (1-f_1(g_1(\bz))) \\
+ \log f_2(\bar{\bx})+ \log (1-f_2(g_2(\bz_{\bar{\bx}})))\\
\mbox{subject to} ~
\boldsymbol \theta_{g_1^{(j)}}=\boldsymbol \theta_{g_2^{(j)}}, ~ j=1,2,\ldots,k\\
\boldsymbol \theta_{f_1^{(n-i)}}=\boldsymbol \theta_{f_2^{(n-i)}},  ~ i=0,1,\ldots,l-1
\end{split}
\end{equation}

The crossing GAN can be interpreted as minimax game with two teams and each team has two players.

\subsection{Implementation Details}

Given a dataset $\bX=\{\bx^{(i)},\bar{\bx}^{(i)}\}_{i=1}^M$ where $N=2M$ is the total number of data points.
\begin{equation}\label{eq:align_batch}
\mathcal{L}_{align}=\frac{1}{M}\sum_{i=1}^M \max(|| \bz_{\bx^{(i)}}-\mbox{ Align}(\bz_{\bar{\bx}^{(i)}}) ||^2, \tau),
\end{equation}

\begin{equation}\label{eq:loss}
L = \mathcal{L}_{vae} + \mathcal{L}_{align} - \mathcal{L}_{gan}
\end{equation}

In this work, we adopt a deep convolutional GAN framework architecture \cite{ConvGAN} and feature matching strategy \cite{ImprovedGAN} for stable and fast-converging training. The visualization of model is shown in Table \ref{tab:network}. Specifically, we use all convolutional nets to replace deterministic spatial pooling functions (such as max pooling) with strided convolutions. This allows the network to learn its own spatial down-sampling. We use this approach in our generator, allowing it to learn its own spatial up-sampling, and discriminator. The overview architecture of Cross-GAN is shown in Table \ref{tab:network}, and the training procedure is summarized in Algorithm \ref{alg:cross-GAN}.

\begin{table*}[tb]\tiny
  \caption{The network architecture of Cross-GANs.}\label{tab:network}
  \hspace{-1cm}
  {
  \begin{tabular}{|l||c|c|c|}
  \hline
  \multirow{2}{*}{Layer} & \multicolumn{3}{c|}{Generator}  \\
  \cline{2-4}
            & View 1 & View 2 &  Shared? \\
  \hline
  1 & Conv (N=20, K=$5\times5$, S=1), BN, ReLU &  Conv (N=20, K=$5\times5$, S=1), BN, ReLU & Yes\\
   2 & Conv (N=20, K=$5\times5$, S=1), BN, ReLU &  Conv (N=20, K=$5\times5$, S=1), BN, ReLU & Yes\\
   3 & Conv (N=20, K=$5\times5$, S=1), BN, ReLU &  Conv (N=20, K=$5\times5$, S=1), BN, ReLU & Yes\\
   4 & Conv (N=20, K=$3\times3$, S=1), BN, ReLU &  Conv (N=20, K=$3\times3$, S=1), BN, ReLU & Yes\\
   5 & Conv (N=20, K=$3\times3$, S=1), BN &  Conv (N=20, K=$3\times3$, S=1), BN & No\\
  \hline
  \hline
   & \multicolumn{3}{c|}{Discriminator}  \\
   \hline
   & View 1 & View 2 &  Shared? \\
   \hline
 1 & Conv (N=20, K=$5\times5$, S=1), MAX-POOL (S=2), LeakyReLU & Conv (N=20, K=$5\times5$, S=1), MAX-POOL (S=2), LeakyReLU  &  No \\
 2 & Conv (N=20, K=$5\times5$, S=1), MAX-POOL (S=2), LeakyReLU & Conv (N=20, K=$5\times5$, S=1), MAX-POOL (S=2), LeakyReLU  &  No \\
 3 & Conv (N=20, K=$5\times5$, S=1), MAX-POOL (S=2), LeakyReLU & Conv (N=20, K=$5\times5$, S=1), MAX-POOL (S=2), LeakyReLU  &  No \\
 4 & FC (N=1024), ReLU & FC (N=1024), ReLU & No \\
 5 & FC (N=1024), Sigmoid & FC (N=1024), Sigmoid & Yes\\
  \hline
  \end{tabular}
  }
\end{table*}

\begin{algorithm}
\SetKwInOut{Input}{input}\SetKwInOut{Output}{output}
\Input{Mini-batch of training samples in pairs $\bX=\{\bx^{(j)},\bar{\bx}^{(j)}\}_{j=1}^M$}
\Output{Parameters of VAE, alignment, and two GANs}
Initialize parameters for VAE and alignment: $\boldsymbol\theta_{vae}$, $\boldsymbol\theta_{align}$\;
Initialize parameters $\boldsymbol \theta_{f_1^{(i)}}$, $\boldsymbol \theta_{f_1^{(i)}}$, $\boldsymbol \theta_{g_1^{(j)}}$, $\boldsymbol \theta_{g_2^{(j)}}$ with the shared network connection weights set to the same values.\;
\For{$t=0,1,2,\ldots,T$}{
\tcc{ update parameters of VAE}
\Repeat{convergence of parameters $\boldsymbol\theta_{vae}$}{
Draw $M$ samples from camera view A, $\{\bx^{(1)},\ldots,\bx^{(M)}\}$\;
Draw $M$ samples from camera view B, $\{\bar{\bx}^{(1)},\ldots,\bar{\bx}^{(M)}\}$\;
$\boldsymbol\epsilon \leftarrow$ random samples from noise distribution $p(\boldsymbol\epsilon)$\;
Compute gradients of the estimator of Eq.\eqref{eq:estimator}: $\boldsymbol e\leftarrow\bigtriangledown_{\boldsymbol\theta_{vae}} \frac{1}{M}\sum_{j=1}^M\left(\tilde{L}_{vae}(\bx^{(j)}) + \tilde{L}_{vae}(\bar{\bx}^{(j)}) \right)$\;
Update parameters of $\boldsymbol\theta_{vae}$ using gradients $\boldsymbol e$ (\eg SGD or Adagrad \cite{Adagrad})
}
\tcc{update parameters of $\boldsymbol\theta_{align}$}
Compute the gradients of the parameters of the alignment  $\bigtriangledown \mathcal{L}_{align}$ (eq.\eqref{eq:align})\;
\tcc{update parameters of two GANs}
Draw $M$ samples from $p(\bz)$, $\{\bz^{(1)},\ldots,\bz^{(M)}\}$\;
Compute the gradients of the parameters of the discriminator, $f_1^t$, $\Delta\boldsymbol\theta_{f_1^{(i)}}$\;
$\bigtriangledown_{\boldsymbol\theta_{f_1^{(i)}}} \frac{1}{M}\sum_{j=1}^M \log f_1^t(\bx^{(j)})+\log(1- f_1^t(g_1^t(\bz^{(j)}_{\bx^{(j)}})))$\;
Compute the gradients of the parameters of the discriminator, $f_2^t$, $\Delta\boldsymbol\theta_{f_2^{(i)}}$\;
$\bigtriangledown_{\boldsymbol\theta_{f_2^{(i)}}} \frac{1}{M}\sum_{j=1}^M \log f_2^t(\bar{\bx}^{(j)})+\log(1- f_2^t(g_2^t(\bz^{(j)}_{\bar{\bx}^{(j)}})))$\;
}
\caption{Mini-batch stochastic gradient descent for training crossing generative adversarial nets.}\label{alg:cross-GAN}
\end{algorithm}

\section{Experiments}\label{sec:exp}

\begin{figure}[t]
\centering
\includegraphics[height=3cm]{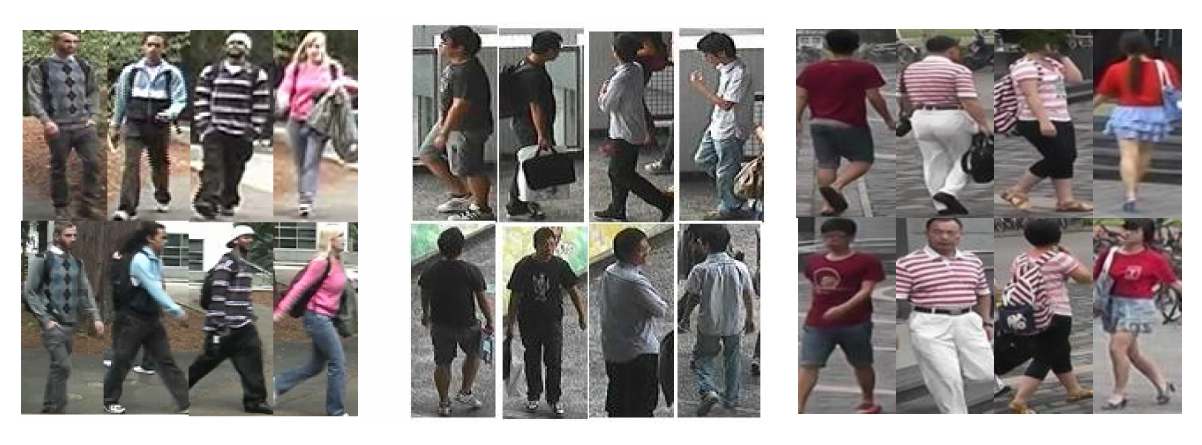}
\caption{Examples from person re-identification datasets: VIPeR (left), CUHK03 (middle), and Market-1501 (right). Columns indicate the same identities.}
\label{fig:dataset}
\end{figure}

\subsection{Datasets and Settings}
We perform experiments on three benchmarks: VIPeR  \cite{Gray2007Evaluating}, CUHK03 \cite{FPNN}, and Market-1501 data set \cite{Market1501}.

\begin{itemize}
\item The \textbf{VIPeR} data set \cite{Gray2007Evaluating} contains $632$ individuals taken from two cameras with arbitrary viewpoints and varying illumination conditions. The 632 person's images are randomly divided into two equal halves, one for training and the other for testing.

\item The \textbf{CUHK03} data set \cite{FPNN} includes 13,164 images of 1360 pedestrians. The whole dataset is captured with six surveillance camera. Each identity is observed by two disjoint camera views, yielding an average 4.8 images in each view. This dataset provides both manually labeled pedestrian bounding boxes and bounding boxes automatically obtained by running a pedestrian detector \cite{DetectionPAMI}. In our experiment, we report results on labeled data set. The dataset is randomly partitioned into training, validation, and test with 1160, 100, and 100 identities, respectively.

\item The \textbf{Market-1501} data set \cite{Market1501} contains 32,643 fully annotated boxes of 1501 pedestrians, making it the largest person re-id dataset to date. Each identity is captured by at most six cameras and boxes of person are obtained by running a detector of Deformable Part Model (DPM) \cite{MarketDetector}. The dataset is randomly divided into training and testing sets, containing 750 and 751  identities, respectively.
\end{itemize}

We use the deep convolutional networks to instantiate the GANs in Cross-GAN. The two generative models have an identical structure with 5 convolutional layers. The generator is realized using the convolutions of ResNet-50 \cite{ResNet} with fine-tuned parameters on re-id \cite{PIE-reid}. Following \cite{CoGAN}, we use the batch normalization processing and the parameter sharing is applied on all convolutional layers except the last convolution. For the discriminative models, we use three fully connected layers with hidden units of 1,024 on each layer. The inputs to the discriminative models are batches containing the output images from the generators and images from each training subsets. Also, each training set is equally divided into two non-overlapping subsets, which are used to train two GANs respectively. The Adam algorithm \cite{ADAM} is used for training, the learning rate is set to be 0.002, the momentum parameter is 0.5, and the mini-batch size is 128. The training is performed 30,000 iterations.

The evaluation protocol we adopt is the widely used single-shot modality to allow extensive comparison. Each probe image is matched against the gallery set, and the rank of the true match is obtained. The rank-$k$ recognition rate is the expectation of the matches at rank $k$, and the cumulative values of the recognition rate at all ranks are recorded as the one-trial Cumulative Matching Characteristic (CMC) results. This evaluation is performed ten times, and the average CMC results are reported.

\subsection{Ablation Studies}\label{ssec:ablation}

\subsubsection{The Impact of Cross-View Alignment}
\begin{figure}[t]
\centering
\includegraphics[height=5cm]{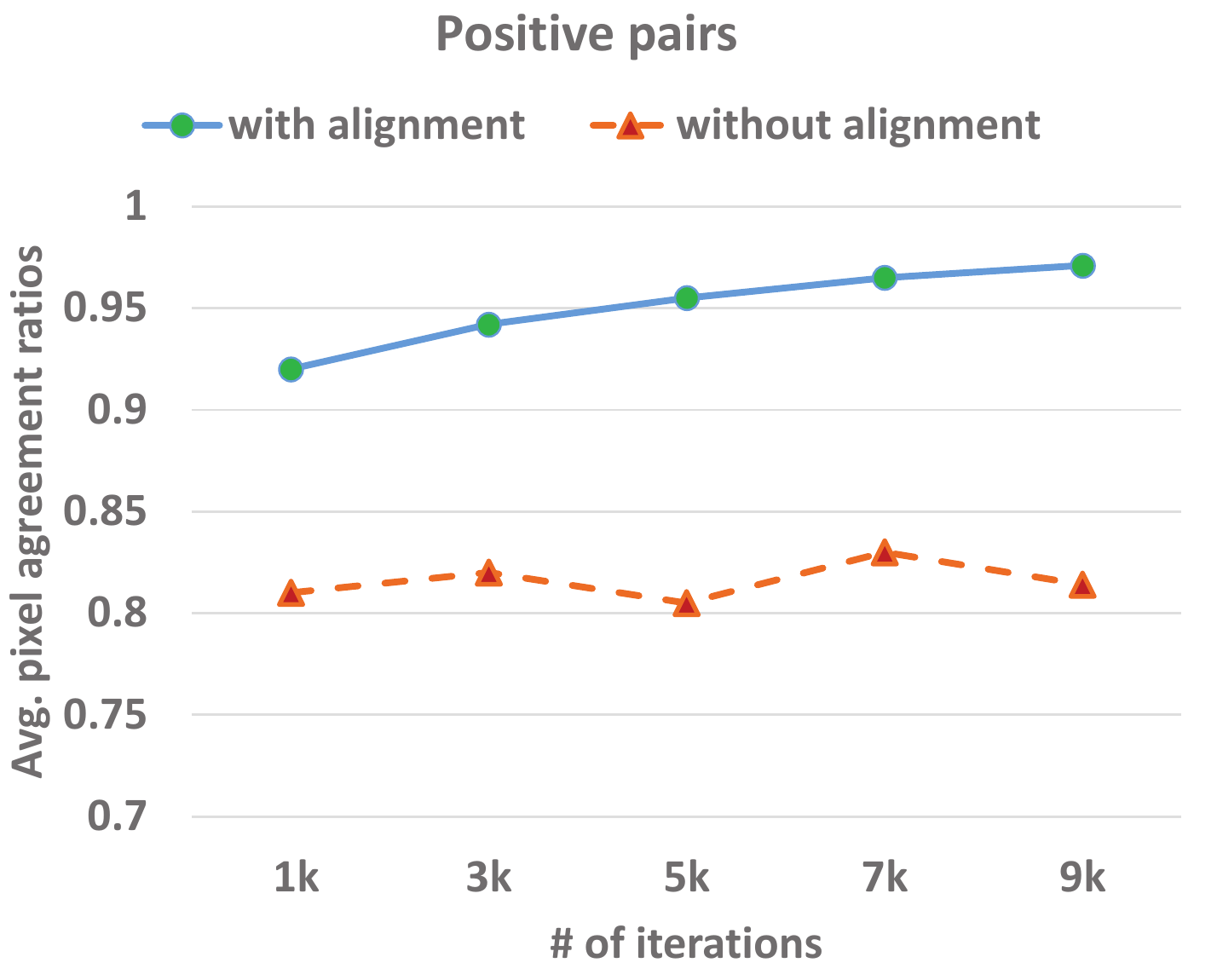}\\
\includegraphics[height=5cm]{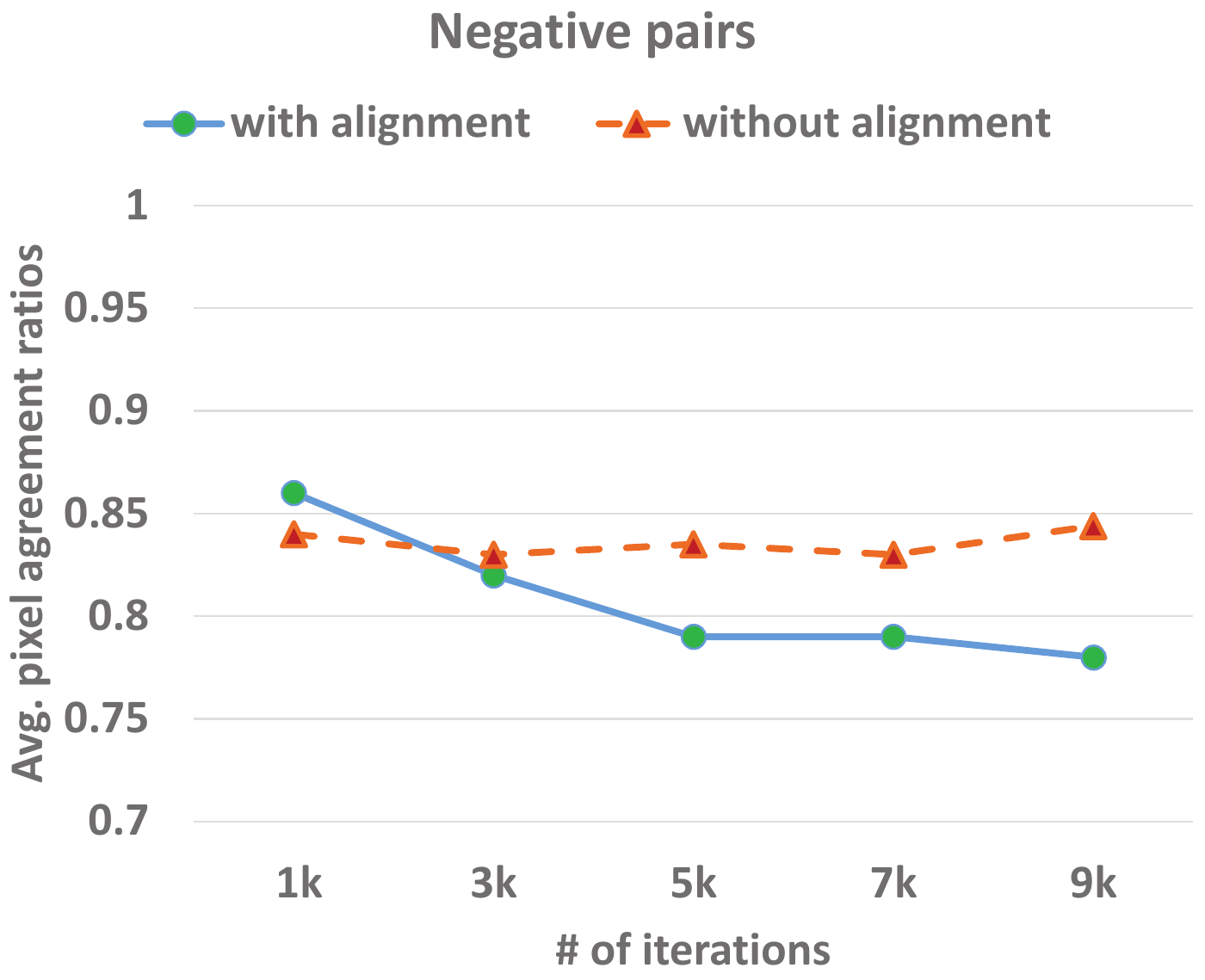}
\caption{The average agreement ratios of the Cross-GAN with/without cross-view alignment on VIPeR dataset.}
\label{fig:avg-agree-ratio}
\end{figure}

In this experiment, we study the impact of cross-view alignment which is demonstrated to be essential to the person matching. To quantifying the performance with/without cross-view alignment, we transform the query images generated by $g_1$ to the gallery view by using the same method employed for generating the training image in the gallery camera view. Then we compare the transformed images with the images generated by the $g_2$. The performance is measured by the average of the ratios of agreed pixels between the transformed image and the corresponding image in the gallery view. The pixel agreement ratio is the number of corresponding pixels that have the same value in the two images divided by the total image size. The experimental results are shown in Fig.\ref{fig:avg-agree-ratio}, and it can be observed that with the cross-view alignment strategy, the rendered pairs of images (positive or negative) resembled true pairs drawn from the joint distributions.

\subsubsection{The Impact of Weight-Sharing in GAN}

\begin{figure}[t]
\centering
\includegraphics[height=5cm]{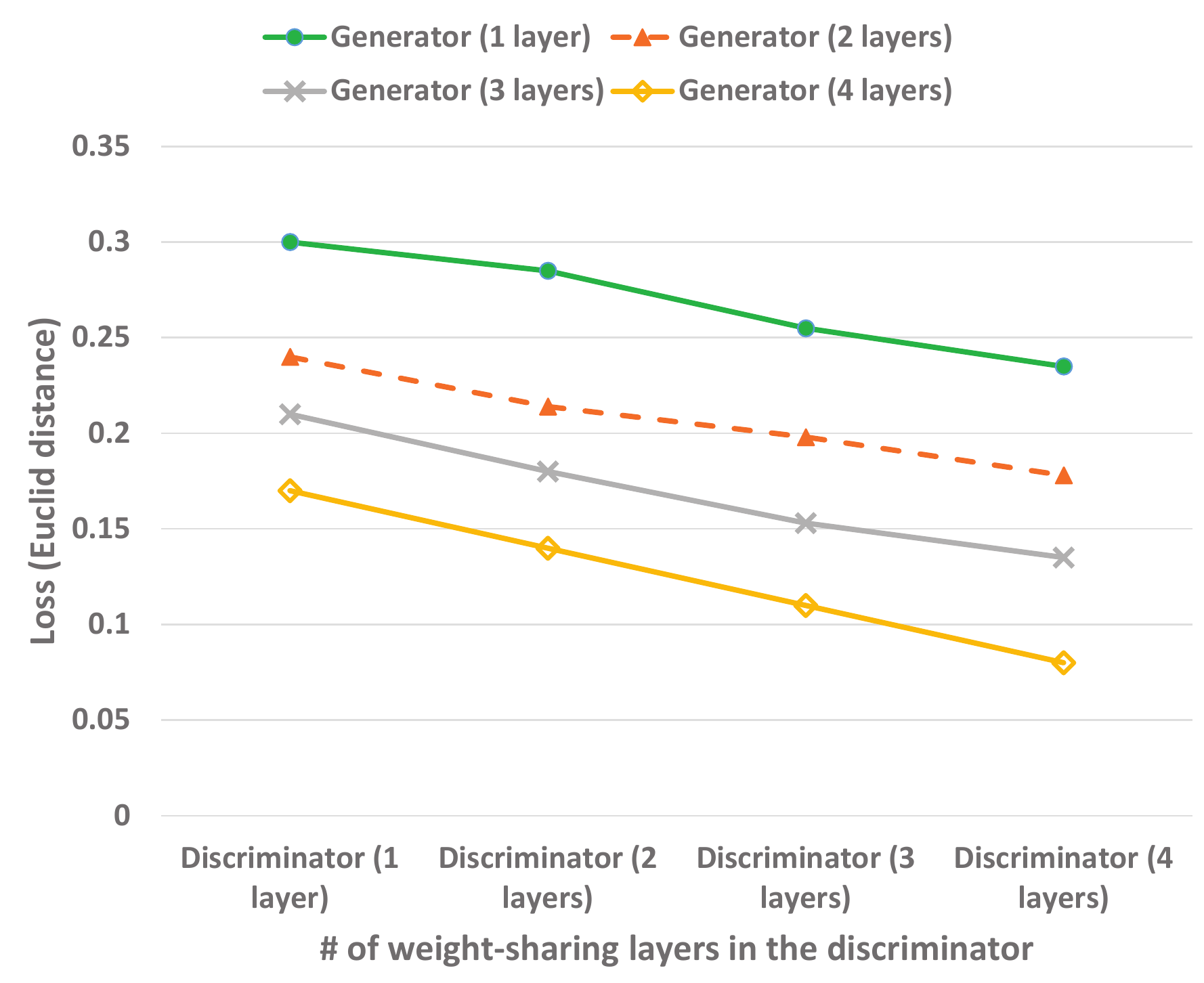}
\caption{The loss function (Euclidean distance) measuring the difference between two images from VIPeR with respect to different weight-sharing configurations in the coupled generators and discriminators. It can be seen that the performance is positively correlated with the number of weight-sharing layers in the generative models but less correlated with the number of weight-sharing layers in the discriminators.}
\label{fig:loss-weight-sharing}
\end{figure}

The weight-sharing constraint and adversarial learning are crucial for co-occurrence pattern encoding/generation across images without requirement on the labeled pair in correspondence. In our model, each sample can be separately drawn from the marginal distribution $p_{\bx_1}$ and $p_{\bx_2}$, and not rely on samples in correspondence with joint distribution of $p_{\bx_1,\bx_2}$. The adversarial learning encourages the generators to produce realistic images individually resembling to respective view domains, while the weight-sharing can capture the correspondence between two views automatically.

In this experiment, we study the weight-sharing effect for the adversarial training by varying the number of weight-sharing layers in both generative and discriminative models. If image $x_1$ is from the probe view, and the Cross-GAN is trained to find the correct matching image $\bar{x}_1$ from the gallery view such that the joint probability density $p(x_1,\bar{x}_1)$ is maximized. Let $L$ be the loss function measuring the difference between two images, e.g., $L$ is implemented to be the Euclidean distance in this experiment. Given $g_1$ and $g_2$, we aim to seek the transformation by finding the random vector that generates the query image via $z^{\ast}=\arg\min_{z} L(g_1(z),x_1)$. With $z^{\ast}$ found, one can apply $g_2$ to produce the transformed image $\bar{x}_1=g_2(z^{\ast})$. In Fig.\ref{fig:loss-weight-sharing}, we show the loss computed on cross-image transformation matching by using Euclidean distance on VIPeR with varied weight-sharing configurations. It can be seen that the matching performance is positively correlated with the number of weight-sharing layers in the generative models, while less correlated with the number of weight-sharing layers in the discriminative models.

\begin{table}[!hbt]\small
  \centering
    \caption{Comparison results with state-of-the-arts on the VIPeR dataset (test person =316).}\label{tab:cmc-viper}
  \begin{tabular}{|l|l|c|c|c|}
    \hline
     \multicolumn{2}{|c|}{\cellcolor{gray} \textcolor{white} {Method}} & \cellcolor{gray} \textcolor{white} {R=1} & \cellcolor{gray} \textcolor{white} {R=10} & \cellcolor{gray} \textcolor{white} {R=20} \\
     \hline
   \parbox[t]{1mm}{\multirow{9}{*}{\rotatebox[origin=c]{90}{Semi/un-supervised}}} & Cross-GAN & \color{red}$\mathbf{49.28}$ &  \color{red}$\mathbf{91.66}$ & \color{red}$\mathbf{93.47}$ \\
   \cline{2-5}
   & LADF \cite{LADF}  & 29.34 & 75.98 & 88.10\\
   & SDALF \cite{Farenzena2010Person} & 19.87  & 49.37 & 65.73\\
   & eSDC \cite{eSDC} & 26.31  & 58.86 & 72.77\\
   & t-LRDC \cite{t-LRDC} &27.40 & 46.00 & 75.10\\
   & OSML \cite{One-shot-RE-ID} & 34.30 & -& -\\
   & CAMEL \cite{CAMEL} & 30.90 & 52.00& 72.50\\
   & OL-MANS \cite{OL-MANS} & 44.90 & 74.40 & 93.60 \\
   & SalMatch \cite{Zhao2013SalMatch} & 30.16  & 62.50 & 75.60 \\
   \hline
   \parbox[t]{1mm}{\multirow{13}{*}{\rotatebox[origin=c]{90}{Supervised}}}
   & MLF \cite{MidLevelFilter} & 29.11 & 65.20 & 79.90\\
   & LocallyAligned \cite{LocallyAlligned} & 29.60 & 69.30 & 86.70\\
   & JointRe-id \cite{JointRe-id} & 34.80  & 74.79 & 82.45  \\
   & SCSP \cite{SimilaritySpatial} & 53.54  & 91.49 &  $\mathbf{96.65}$\\
   & Multi-channel \cite{Multi-channel-part} & 47.80  & 84.80 & 91.10\\
   & DNSL \cite{NullSpace-Reid} & 42.28  & 82.94 & 92.06\\
   & JSTL \cite{DomainDropout} &38.40 & - & -\\
   & SI-CI \cite{SI-CI} & 35.80 & 83.50 & -\\
   & S-LSTM \cite{S-LSTM} &42.40 & 79.40 &-\\
   & S-CNN \cite{GatedCNN} &37.80 & 77.40&-\\
   & SpindleNet \cite{SpindleNet} & \color{blue}$\mathbf{53.80}$ & 90.10 & 96.10\\
   & Part-Aligned \cite{Part-Aligned} & 48.70 & 87.70 & 93.00\\
   & Deep-Embed \cite{Deep-Embed} & 49.00  & 91.10  &96.20 \\
    \hline
  \end{tabular}
\end{table}

\begin{table}[!hbt]\small
  \centering
  \caption{Rank-1, -10, -20 recognition rate of various methods on the CUHK03 data set (test person =100).}  \label{tab:cmc_cuhk03}
  {
  \begin{tabular}{|l|l|c|c|c|}
    \hline
     \multicolumn{2}{|c|}{\cellcolor{gray} \textcolor{white} {Method}} & \cellcolor{gray} \textcolor{white} {R=1} & \cellcolor{gray} \textcolor{white} {R=10} & \cellcolor{gray} \textcolor{white} {R=20} \\
     \hline
   \parbox[t]{1mm}{\multirow{8}{*}{\rotatebox[origin=c]{90}{Semi/un-supervised}}} & Cross-GAN & \color{red}$\mathbf{83.23}$ &  \color{red}$\mathbf{96.73}$ & \color{red}$\mathbf{99.47}$ \\
   \cline{2-5}
   & OSML \cite{One-shot-RE-ID} & 45.61 & 85.43 & 88.50 \\
   & LSRO \cite{LSRO} & 84.62 & 97.64 & 99.80 \\
   & CAMEL \cite{CAMEL} &31.90 & 76.62 & 80.63 \\
   & eSDC \cite{eSDC} & $8.76$  &38.28 & 53.44 \\
   & UMDL \cite{UMDL} & 1.64 & 8.43 & 10.24 \\
   & OL-MANS \cite{OL-MANS} & 61.70 & 92.40 & 98.52 \\
   & XQDA \cite{LOMOMetric} & 52.20  & 92.14 & 96.25\\
   \hline
   \parbox[t]{1mm}{\multirow{10}{*}{\rotatebox[origin=c]{90}{Supervised}}}
   & FPNN \cite{FPNN} & $20.65$ & 51.32  & 83.06\\
   & kLFDA \cite{Xiong2014Person} & 48.20  & 66.38 & 76.59\\
   & DNSL \cite{NullSpace-Reid}  & $58.90$  & $92.45$ & 96.30 \\
   & JointRe-id \cite{JointRe-id}  & $54.74$  & 91.50 & 97.31 \\
   & E-Metric \cite{E-Metric} & 61.32 & 96.50 & 97.50 \\
   & S-LSTM \cite{S-LSTM} & 57.30 & 88.30& -\\
   & S-CNN \cite{GatedCNN} & 61.80 & 88.30 & -\\
   & Deep-Embed \cite{Deep-Embed} & 73.00 & 94.60 & 98.60\\
   & SpindleNet \cite{SpindleNet} & \color{blue}$\mathbf{88.50}$ & 98.80 & 99.20\\
   & Part-Aligned \cite{Part-Aligned} & 85.40 & 98.60 & 99.90\\
\hline
  \end{tabular}
  }
\end{table}

\begin{table}[!hbt]\tiny
  \centering
  \caption{Rank-1, -10, -20 recognition rate and mAP of various methods on the Market-1501 data set (test person =751). All results are evaluated on single-shot setting.}\label{tab:cmc_market}
  {
  \begin{tabular}{|l|c|c|c|c|c|}
   \multicolumn{2}{|c|}{\cellcolor{gray} \textcolor{white} {Method}} & \cellcolor{gray} \textcolor{white} {R=1} & \cellcolor{gray} \textcolor{white} {R=10} & \cellcolor{gray} \textcolor{white} {R=20} & \cellcolor{gray} \textcolor{white} {mAP}\\
     \hline
   \parbox[t]{1mm}{\multirow{10}{*}{\rotatebox[origin=c]{90}{Semi/un-supervised}}} & Cross-GAN & \color{red}$\mathbf{72.15}$ &  \color{red}$\mathbf{94.3}$ & \color{red}$\mathbf{97.5}$ & \color{red} $\mathbf{48.24}$\\
   \cline{2-6}
   & eSDC \cite{eSDC} & 33.54 & 60.61 & 67.53 & 13.54\\
   & SDALF \cite{Farenzena2010Person} & 20.53 & - & - & 8.20\\
   & LSRO \cite{LSRO} & \color{blue}$\mathbf{83.97}$ & 95.64 & 97.56 & 66.07\\
   & CAMEL \cite{CAMEL} & 54.56 &  84.67 & 87.03 & -\\
   & OL-MANS \cite{OL-MANS} &60.72 & 89.80& 91.87 &-\\
   & PUL \cite{PUL} & 45.53 & 72.75 & 72.65 &-\\
   & UMDL \cite{UMDL} & 34.54 & 62.60 & 68.03 &-\\
   & XQDA \cite{LOMOMetric} & 43.79 & 75.32 & 80.41 & 22.22\\
   & BoW \cite{Market1501} & 34.40 & -& -& 14.09\\
    \hline
   \parbox[t]{1mm}{\multirow{9}{*}{\rotatebox[origin=c]{90}{Supervised}}} & JSTL \cite{DomainDropout} & 44.72 & 77.24 & 82.00 & -\\
   & KISSME \cite{KISSME} & 39.35 & -&- & 19.12\\
   & kLFDA \cite{Xiong2014Person} & 44.37 &- &- & 23.14 \\
   & SCSP \cite{SimilaritySpatial} & 51.90 &- &- & 26.35\\
   & DNSL \cite{NullSpace-Reid}  & 61.02 &- & -& 35.68\\
   & S-CNN \cite{GatedCNN} & 65.88 & - & -& 39.55\\
   & Deep-Embed \cite{Deep-Embed} & 68.32 & 94.59 & 96.71 & 40.24 \\
   & SpindleNet \cite{SpindleNet} & 76.90 & - & - & -\\
   & Part-Aligned \cite{Part-Aligned} & 81.00 & - & - &-\\
\hline
  \end{tabular}
  }
\end{table}

\subsection{Comparison with State-of-the-arts}

In this subsection, we extensively compare the proposed Cross-GAN with a number of state-of-the-art semi/unsupervised and supervised methods on three datasets. Semi/unsupervised methods include LADF \cite{LADF}, SDALF \cite{Farenzena2010Person}, eSDC \cite{eSDC}, t-LRDC \cite{t-LRDC}, OSML \cite{One-shot-RE-ID}, CAMEL \cite{CAMEL}, OL-MANS \cite{OL-MANS}, SalMatch \cite{Zhao2013SalMatch}, UMDL \cite{UMDL}, XQDA \cite{LOMOMetric}, and PUL \cite{PUL}. Supervised methods include MLF \cite{MidLevelFilter}, LocallyAligned \cite{LocallyAlligned}, JointRe-id \cite{JointRe-id}, SCSP \cite{SimilaritySpatial}, Multi-channel \cite{Multi-channel-part}, DNSL \cite{NullSpace-Reid}, JSTL \cite{DomainDropout}, SI-CI \cite{SI-CI}, S-CNN \cite{GatedCNN}, SpindleNet \cite{SpindleNet}, Part-Aligned \cite{Part-Aligned}, FPNN \cite{FPNN}, S-LSTM \cite{S-LSTM}, kLFDA \cite{Xiong2014Person}, KISSME \cite{KISSME}, E-Metric \cite{E-Metric} and Deep-Embed \cite{Deep-Embed}. Please note that not all methods report their matching results on three datasets and the CMC values are quoted from their papers.

\begin{figure*}[t]
\centering
\hspace{-2cm}
\begin{tabular}{ccc}
\includegraphics[width=5cm,height=3.5cm]{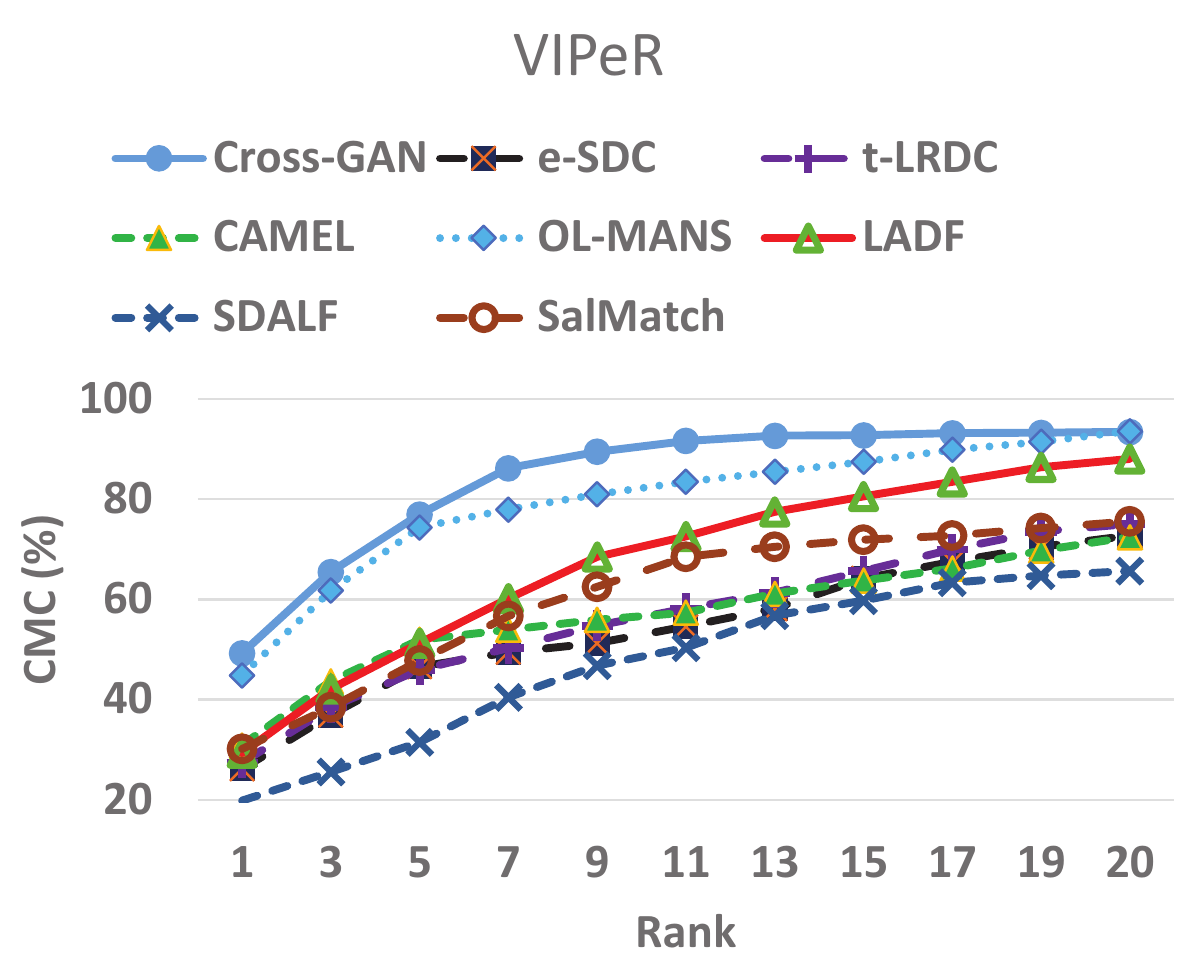} &
\includegraphics[width=5cm,height=3.5cm]{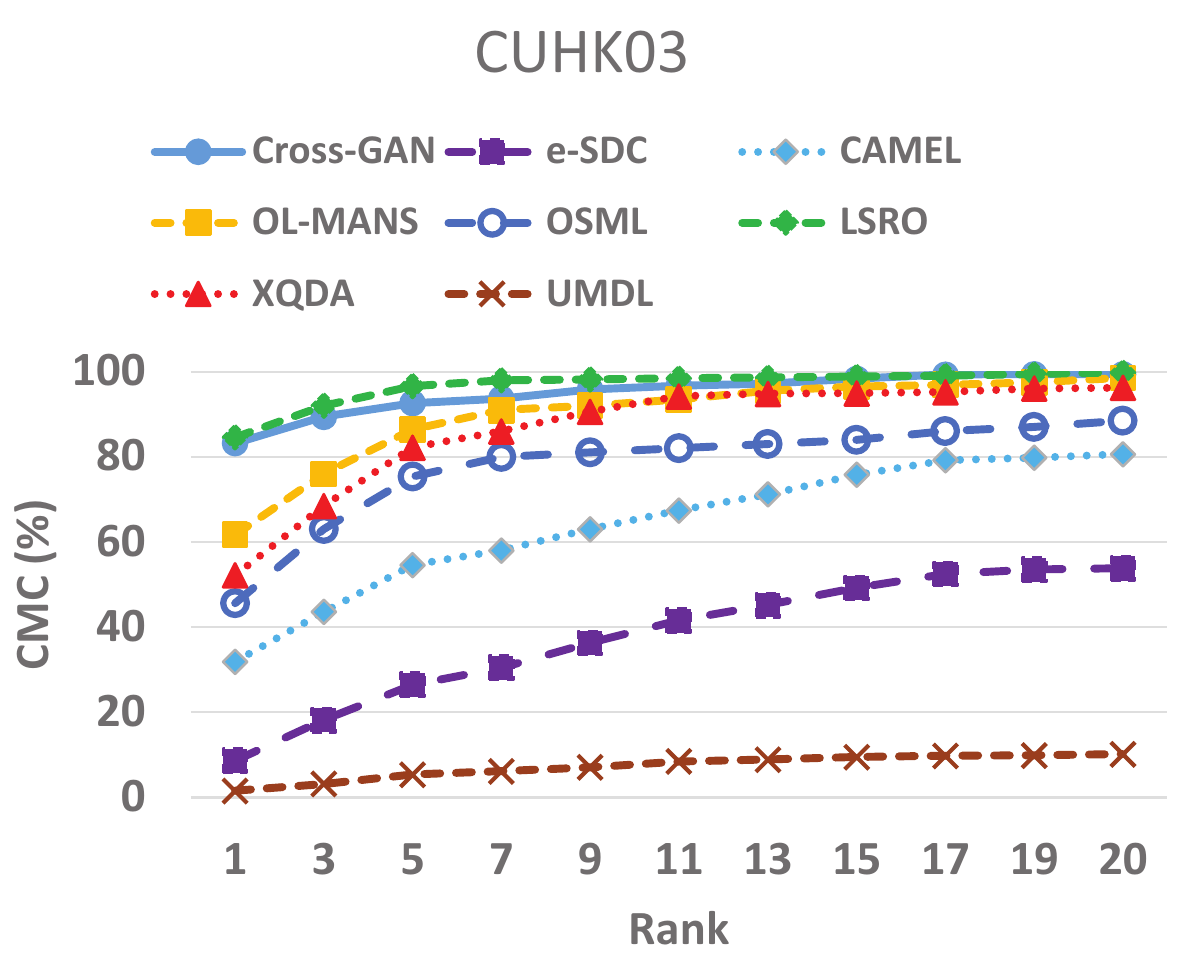} &
\includegraphics[width=5cm,height=3.5cm]{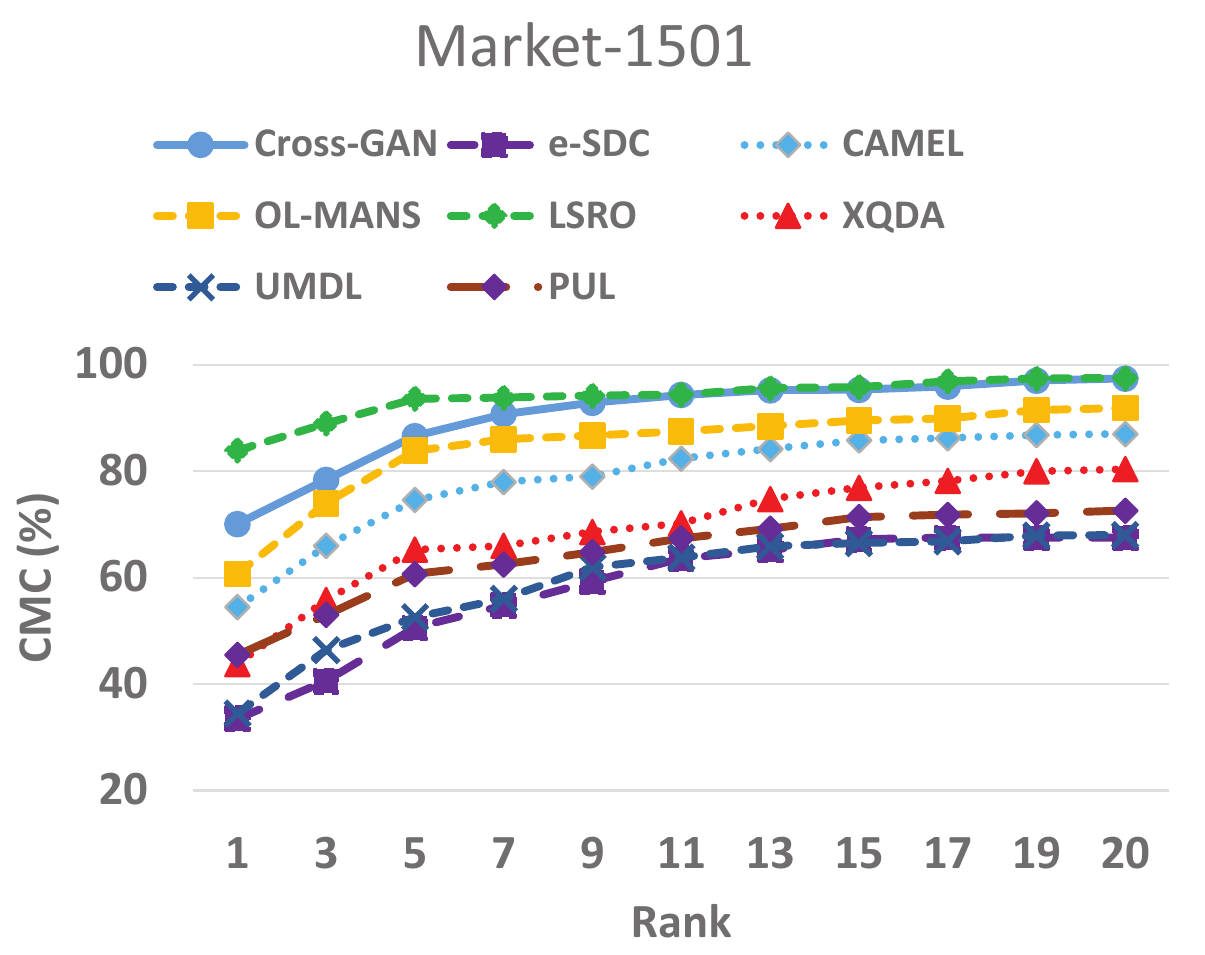}\\
(a) VIPeR & (b) CUHK03 & (c) Market-1501
\end{tabular}
\caption{CMC curves of unsupervised/semi-supervised methods on three datasets.}\label{fig:cmc}
\end{figure*}

The comparison results are reported in Table \ref{tab:cmc-viper}, Table \ref{tab:cmc_cuhk03}, and Table \ref{tab:cmc_market} for VIPeR, CUHK03, and Market-1501 respectively. The CMC curves of unsupervised/semi-supervised methods on three datasets are shown in Fig.\ref{fig:cmc}.

In the VIPeR dataset, Cross-GAN notably outperforms all semi/unsupervised competitors by achieving rank-1=49.28. Compared with unsupervised feature encoding methods such as SDALF \cite{Farenzena2010Person}, eSDC \cite{eSDC}, and SalMatch \cite{Zhao2013SalMatch}, the proposed method of Cross-GAN is able to learn deep local features with joint distribution and thus robust against visual variations. Also, our method is very comparable to the state-of-the-art supervised method of SpindleNet \cite{SpindleNet} which obtains rank-1=53.80.

In the CUHK03 dataset, the proposed Cross-GAN outperforms all state-of-the-art unsupervised method except LSRO \cite{LSRO} whereby Cross-GAN achieves rank-1=83.23 versus LSRO \cite{LSRO} achieves rank-1=84.62. The main reason is that LSRO \cite{LSRO} is a semi-supervised approach which uses GANs to generate complex realistic images to augment the number of training data, and a uniform labeling on the generated samples and semantic labeling on existing training samples are performed respectively. However, the proposed method doesn't require any labeling in training.

In the Market-1501 dataset, the matching rate of Cross-GAN is only secondary to LSRO \cite{LSRO}. The primary reason is that on Market-1501, many persons exhibit similar visual appearance and it is more difficult to distinguish people without any supervision aid. In this aspect, LSRO \cite{LSRO} generates more realistic images regarding each person to enable discriminative feature learning. However, generating sophisticated images in large numbers is very computationally expensive, which is not feasible in practice. In contrast, the proposed Cross-GAN can still achieve very comparable performance to LSRO \cite{LSRO} without any labeling.

\section{Conclusions and Future Work}\label{sec:con}

This paper presents a unsupervised generative model to learn jointly invariant features for person re-id without relying on labeled image pairs in correspondence. The proposed method is built atop variational auto-encoders, a cross-view alignment, and dual GANs to seek a series of non-linear transformations into a shared latent space which allows comparable matching across camera views. The learned joint feature distribution effectively captures the co-occurrence patterns in person image against dramatic visual variations. Extensive experiments are conducted to demonstrate the effectiveness of our method in person re-id by setting the state-of-the-art performance.

\bibliographystyle{named}
\bibliography{ijcai16}

\end{document}